\documentclass[letterpaper]{article} 
\usepackage{aaai24}  
\usepackage{times}  
\usepackage{helvet}  
\usepackage{courier}  
\usepackage[hyphens]{url}  
\usepackage{graphicx} 
\usepackage{natbib}  
\usepackage{caption} 
\usepackage{algorithm}
\usepackage{algorithmic}
\usepackage{amsfonts,amssymb,amsthm}
\usepackage{etoolbox}
\usepackage[T1]{fontenc}

\usepackage{tikz}
\newcommand*\circled[1]{\tikz[baseline=(char.base)]{
            \node[shape=circle,draw,inner sep=1pt, minimum size=1.5em] (char) {#1};}}

\newcommand{\HEF}[1]{\ifstrempty{#1}{\textrm{\textup{HEF}}}{\textrm{\textup{HEF-{$#1$}}}}}

\newtheorem{example}{Example}

\usepackage{newfloat}
\usepackage{listings}
\DeclareCaptionStyle{ruled}{labelfont=normalfont,labelsep=colon,strut=off} 
\floatstyle{ruled}
\newfloat{listing}{tb}{lst}{}
\floatname{listing}{Listing}
\title{
The Fairness Fair: Bringing Human Perception into\\ Collective Decision-Making
}

\author {
    Hadi Hosseini
}
\affiliations{
    Pennsylvania State University\\
    University Park, PA, USA\\
    hadi@psu.edu
}

\begin{document}

\maketitle

\begin{abstract}
Fairness is one of the most desirable societal principles in collective decision-making. It has been extensively studied in the past decades for its axiomatic properties and has received substantial attention from the multiagent systems community in recent years for its theoretical and computational aspects in algorithmic decision-making.
However, these studies are often not sufficiently rich to capture the intricacies of human perception of fairness in the ambivalent nature of the real-world problems.
We argue that not only fair solutions should be deemed desirable by social planners (designers), but they should be governed by human and societal cognition, consider perceived outcomes based on human judgement, and be verifiable. We discuss how achieving this goal requires a broad transdisciplinary approach ranging from computing and AI to behavioral economics and human-AI interaction. In doing so, we identify shortcomings and long-term challenges of the current literature of fair division, describe recent efforts in addressing them, and more importantly, highlight a series of open research directions.

\end{abstract}


\section{Introduction}

Fairness is a fundamental societal principle for promoting ethical solutions in algorithmic decision-making.
It has become the center of attention in the age of Internet economics \cite{moulin2019fair,moulin2004fair} that employs collective decision-making to deal with multiple stakeholders. 
Arguably, one of the most notable advancements in multiagent systems is setting policies, norms, and axioms of interactions. 
Recent progress in this arena has called for ethical decision making \cite{murukannaiah2020new} in multiagent applications including algorithmic hiring \cite{schumann2020we}, blockchain technology \cite{grossi2022social}, and in general computational social choice \cite{lang2016fair,ijcai2022p756}.

Within computational social choice, \textit{fair division} has emerged to provide practical solutions for distributing goods, services, or tasks in a wide array of problems that people face in their daily lives. Its primary objective is promoting social values such as fairness and welfare with provable axiomatic guarantees.
Fair division has provided a rich framework for studying mathematical foundations of societal good and has enabled advancements in algorithmic solutions that adhere to fairness concepts such as envy-freeness.
A number of fair algorithms, particularly in allocation of \textit{indivisible} items, have been made available to the public for real-world decisions through efforts such as Spliddit (\url{www.spliddit.org}), MatchU.ai (\url{www.matchu.ai}), and Course Match (University of Pennsylvania).

Over the past decades, a plethora of approximate fairness axioms have been proposed to escape negative results driven by i) impossibilities of guaranteeing certain fairness notions, ii) computational intractability, and iii) incompatibility with other socially desirable notions.
Despite the rigorous research in this area, to date, the choice of which fairness axiom (or approximations) to choose has remained largely unexplored, leaving social planners uncertain as to which algorithmic solutions to adopt in practice.
Furthermore, the advent of social platforms, online market places, and participatory resource allocation \cite{lee2019webuildai} has shifted its attention from focusing on agents powered by artificial intelligence (AI) to human-human or human-AI interactions \cite{endriss2017trends,brandt2016handbook}.
Thus, recent efforts in collective decision-making \cite{d2020testing,Clippel22:Cooperative,nizri2022improving} have focused on empirical validation of axioms in computational social choice.
The standard theoretical and algorithmic models of fair division often do not capture the intricacies of real-world settings that are shaped by individuals' cognitive abilities, the availability of information, and more importantly, the community and human psyche. Hence, a fundamental question emerges as to which  axioms of fairness (or relaxation thereof) are perceived to be fairer? And why?

\begin{figure}
\begin{center}
\includegraphics[width=.45\textwidth]{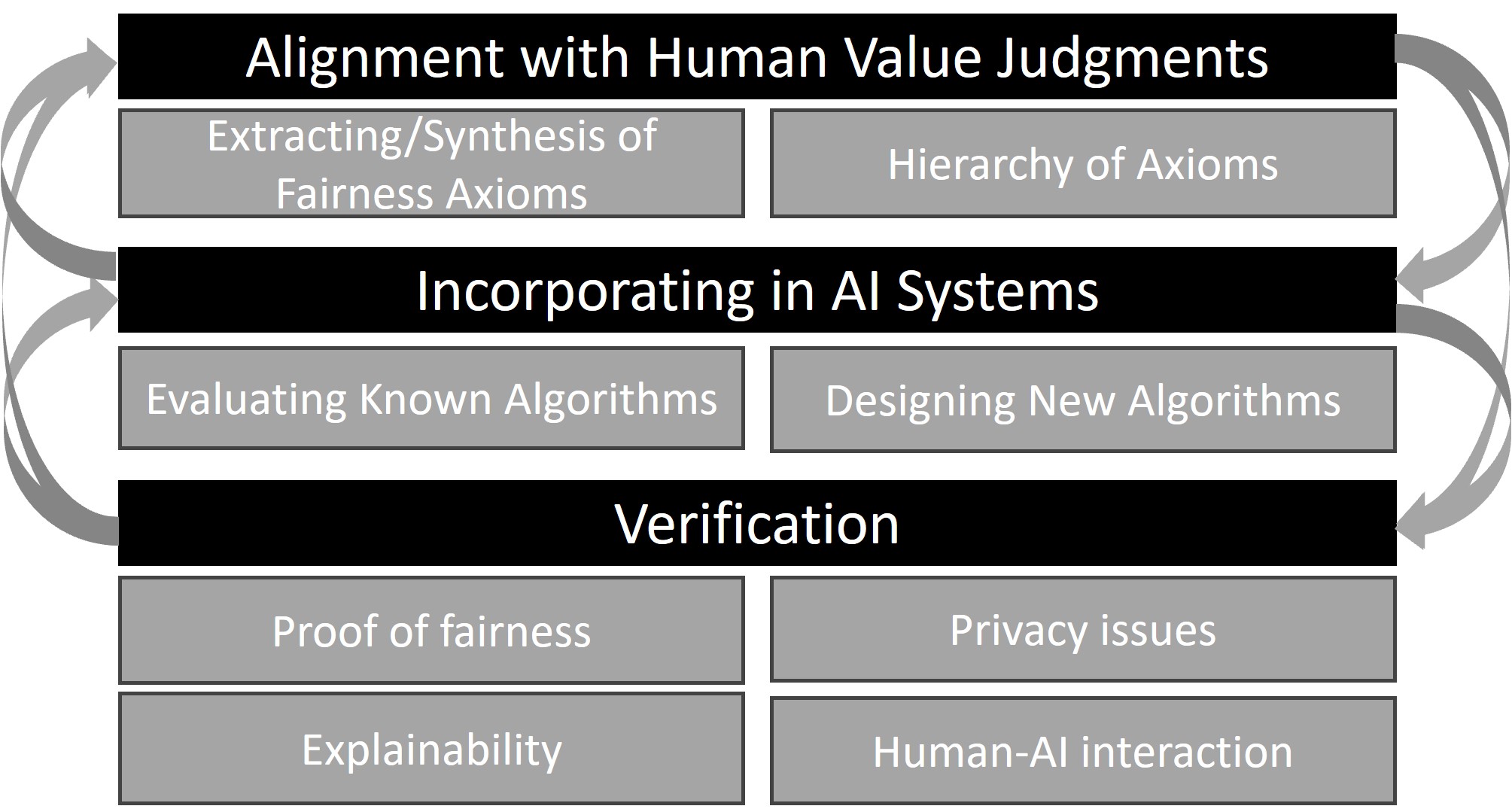}
\end{center}
\caption{A high-level steps in the fairness axiom lifecycle.}
\label{fig:fig1}
\end{figure}

\paragraph{Overview.}
In this paper, we argue for a broader agenda in fair division based on perceived fairness, one that grounds fairness judgements based on human values.\footnote{These new directions may be applied to a broader set of problems in computational social choice (e.g. in voting or cake-cutting). Here, we primarily focus on fair allocation of indivisible resources for the ease of exposition.}
With the popularity of axiomatic approach in online systems \cite{tennenholtz_zohar_moulin_2016}, \citet{procaccia2019axioms} articulates---through two examples rooted in fair division and voting---that axioms should explain solutions beyond providing mathematical guarantees. 
We further postulate that not only axioms of fairness should explain solutions, but they should be 
1) aligned with societal and cognitive values, 
2) incorporated in AI systems that are acceptable by stakeholders (e.g. individuals, communities) and social planners (and policy makers), and 
3) be verifiable, i.e. individuals (or collectives) should be able to verify the solutions (see Figure~\ref{fig:fig1}).

Bringing human perception into decision-making leads to furthering our understanding of societal fairness when deploying algorithms based on various fair solutions, and ultimately leads to the development of new axioms of fairness.
Understanding what is fair and the perception of fairness for individuals or a society of agents requires a trans-disciplinary approach from AI and computing research to experimental economics and behavioral psychology. Focusing on a broad perspective, we discuss challenges in incorporating human perception along the dimensions of \textit{value judgements}, \textit{integration}, and \textit{verification}.
Thus, the overarching questions are:
\begin{quote}
    \textit{What axioms are more aligned with human value judgments among all known fairness axioms?
    What (and how) cognitive and behavioral factors influence individuals' perception of fairness, and how should these human judgements form new fairness concepts and inform the design of new algorithms?}
\end{quote}

\section{The Fairness Fair}

A standard fair division problem aims at distributing a set of $m$ indivisible resources (or goods) among a set of $n$ agents such that the resulting solution satisfies some axioms of fairness. The preference of an agent is generally specified with a valuation function $v_i$ that maps each `solution' to a value. This model allows for a wide variety of preferences over the outcomes beyond the standard assumptions in fair division that often consider valuation functions to be (i) \textit{idiosyncratic} with \textit{no externalities} and (ii) \textit{additive} (the utility of each individual is the sum of values of individual items).



The primary fairness notions can be seen as either \textit{comparison-based} notions that rely on pairwise comparisons (e.g. envy-freeness and its relaxations), and \textit{threshold-based} criteria that deal with achieving a fair-share of the set of items (e.g. proportionality and maximin share).\footnote{We do not discuss notions that measure the degree of fairness in a society by minimizing the maximum or sum of envy \cite{chevaleyre2007reaching,chen2017ignorance,nguyen2014minimizing}, minimizing the envy ratio \cite{LMM+04approximately}, or balancing the amount of envy experienced in a society.}

One of the most desirable comparison-based axiom of fairness is \textit{envy-freeness} (EF), which requires that no agent prefers the allocation of another agent to its own \citep{foley1967resource,george1958puzzle}. 
A well-studied threshold-based axiom is \textit{proportionality} that requires giving each agent to receive at least $1/n$ of its value for all the items. Its extension, maximin share (MMS), is the value an agent can guarantee by partitioning the items into $n$ bundles and receiving the least preferred one \cite{B11combinatorial}.
It is crucial to note that fairness is often studied in conjunction with economic efficiency notions such as Pareto optimality to guarantee some level of social welfare. 

\subsubsection{Relaxations.}
When dealing with indivisible items, none of the above fairness notions can be guaranteed. For example, consider a single valuable item and two agents. Moreover, given an instance, deciding whether such an allocation exists is generally proven to be computationally intractable. 
These negative results have motivated deterministic relaxations of such solution concepts (in Section \ref{sec:uncertainty} we discuss an alternative method that utilizes randomization).


The most prominent relaxations of fairness axioms in the literature are envy-freeness up to one item (EF1) \cite{LMM+04approximately}, envy-freeness up to any item (EFX) \cite{CKM+19unreasonable}, and (approximate) maximin share fairness (MMS) \cite{B11combinatorial}. While EF1 allocations always exist and can be computed in polynomial time \cite{LMM+04approximately,CKM+19unreasonable}, the existence and computation of EFX allocations----with some exceptions for instance when $n<4$ or under lexicographic extensions \cite{CGM20efx,hosseini2021lexpreferences}---are intriguing open problems that have motivated a large array of works in recent years. Similarly, the non-existence of MMS allocations \cite{KPW18fair} has inspired a variety of algorithmic techniques in finding multiplicative \cite{ghodsi2018fair,garg2020improved} or ordinal \cite{hosseini2022ordinal,hosseini2022ordinalchores} approximations to the MMS threshold.\footnote{For a comprehensive list, see the following recent surveys \cite{ALM+22algorithmic,ABF+22fair}.}
The following example illustrates an instance of a fair division problem. We will use it as a running example when discussing the challenges and directions in incorporating human perception in fair decision-making.

\begin{example}\label{ex:runningexample}
    Consider three agents with preferences as shown in Table \ref{tab:example}. For simplicity of exposition, the preferences are assumed to be additive.
    The allocation shown with circles is EF1, but does not satisfy proportionality: Agent 1 is envy-free; the envy of agents 2 and 3 towards agent 1 can be eliminated by the hypothetical removal of a single item ($g_3$). 
    The allocation shown with underlines is likewise EF1 but improves the economic efficiency (more precisely, it is EF1 and Pareto optimal).
    \begin{table}[t]
\begin{center}\small
\begin{tabular}{c| c c c c }
               & $g_{1}$   & $g_{2}$   & $g_{3}$   & $g_{4}$       \\\hline
        1 & \circled{\underline{6}}         & 4         & \circled{10}         & \underline{10}          \\ 
        2 & 7       & 6         & \underline{9}          & \circled{8}         \\
        3 & 1       & \circled{\underline{11}}         & 12          & 6         \\
    \end{tabular}
\end{center}
\caption{A stylized running example with four items and three agents.}
\label{tab:example}
\end{table}    
\end{example}



\section{Alignment with Human Value Judgement}

In the field of fair division, the most common fairness axioms were originated from the early works of mathematicians such as Steinhaus on fairly dividing a cake \cite{steinhaus1948problem}
and insights that followed from philosophical theories in distributive justice \cite{rawls1971theory,sen2018collective}.
Over the next few decades, these studies resulted in the development of a plethora of fairness axioms.
While these mathematical constructs gave rise to advancements in mechanism design and algorithmic solutions, they do not completely capture the intricacies of human decision judgement, as often fairness axioms (or outcomes) themselves influence the preferences of individuals \cite{sandbu2008axiomatic}.

In what follows, we argue that 
1) the well-studied fairness concepts should be revisited and evaluated in the presence of complexities in human perception,
2) it is critical to investigate competing fairness axioms and their (in)compatibility when interacting with individuals, and
3) there is a dire need for devising a \textit{hierarchy of axioms} in fair decision-making to ensure that the axioms are aligned with human preferences.

\subsection{Extracting Fairness and Hierarchy of Axioms}

A compelling argument for the aforementioned fairness concepts is that they do not rely on \textit{interpersonal} utility comparisons\footnote{For a comprehensive discussion on interpersonal utility comparisons and arguments for and against them in the utility theory, see \cite{sen2018collective,harsanyi1990interpersonal}.}, meaning that they eliminate the need for identifying who derives the most (or the least in case of chores/tasks) from (a bundle of) resources. Thus, an agent values its own bundle \textit{only} according to its own direct \textit{introspective} preferences.
Yet, in reality human value judgement is seldom introspective and is often affected by a variety of individual, social, and cognitive parameters. 
This raises the questions of \textit{which fairness axioms are perceived to be fairer according to human values? Do envy-free allocations (when they exist) are perceived to be fairer compared to their proportional counterparts? What about their relaxations? Which fairness axioms are more aligned with human value judgements?}

\subsection{Perceived Fairness and Cognitive Aspects}

In this section, we highlight a few  (non-exhaustive) factors that are often involved in human judgement and discuss their impacts on algorithmic fairness. 


\subsubsection{Transparency and Information Sharing.} \label{sec:transparency}

Counterfactual reasoning seems to be the cornerstone of several approximate fairness concepts.
Many prominent relaxations of fairness such as EF1 and EFX rely on a counterfactual interpretation of fairness, assuming that an agent's envy is eliminated if a single item is to be removed from the envied agents. While transparency is a crucial societal concern, in practice often agents do not have access to all the information. In this vein, \textit{epistemic} relaxations of fairness have been proposed that rely either on  \textit{withholding} information \citep{ABC+18knowledge} or on removal of a randomly selected item \cite{farhadi2021almost}.
A question is \textit{how to utilize information sharing/withholding to increase the perceived fairness of outcomes?}

A recent epistemic relaxation is \emph{envy-freeness up to $k$ hidden goods} (\HEF{k}) that eliminates envy by hiding a small ($k$) number of items \cite{hosseini2020fair}.
Interestingly, an experimental study showed that \HEF{k} allocations are perceived to be fairer compared to their EF1 counterparts \cite{hosseini2022hide}.
Yet, these studies are limited in scope as they pertain to specific settings with highly stylized instances. 
Moreover, it is not clear how information sharing guide (or cause) an outcome to be perceived fairer. One plausible justification could revolve around individuals' bounded rationality or the amount of effort needed to consider the `unknown', which has been described by Daniel Kahneman as `\textit{what you see is all there is}'---the cognitive phenomenon that human brain is hardwired to believe that the given information is the entirety of the relevant information.

\subsubsection{Fair for Me vs. Fair for All.}

In fair division, agents make judgements over their own allocations, either in comparison with what is allocated to other agents (as in envy-freeness) or by measuring its value against a fairness threshold (as in proportionality). However, in reality fairness judgments are often made based on human values for fair outcomes \textit{for all}, meaning that what is fair for me may not be considered as a fair outcome for everyone.

In the context of \textit{bargaining}, \citet{herreiner2009envy} demonstrated experimentally that the primary factor of fairness is in fact \textit{inequality aversion}---i.e. the tendency of humans in seeking equal outcomes---and that envy-freeness plays a secondary role in the perception of fairness, only when economic efficiency and inequality aversion are satisfied.
In other words, everything equal, people tend to associate a higher order to inequality aversion. These findings suggest that individuals seem to (implicitly) create a hierarchy of axioms to reason about relative value of each outcome.
Hence, developing fair algorithms may require axioms that are able to capture solutions that take the social context into account beyond perceived envy.

Let us illustrate this point by visiting Example \ref{ex:runningexample}: Agent 1 may not perceive any unfairness in the circled allocation as she does not envy others, however, she may perceive the overall outcome to be unfair since she receives a higher value (as well as higher number of items) than the other two agents. Does agent 1 consider the outcome fair for agent 3 since she receives her proportional share?
More broadly, \textit{which fairness concepts are employed in assessing outcomes? And do those fairness concepts change when evaluating an outcome for self vs. for everyone?}



\subsubsection{Skin in the Game.}

When measuring fairness of an outcome for everyone, there is an element of how much stake an agent has in the decision itself. In other words, human judgement is often dependent on whether (and to what extent) they have `skin in the game'. 
Looking into the fairness notions such as envy-freeness, it is crucial to understand the perception of fairness both from the eyes of participating agents and non-participating \textit{bystanders}. 
This concept is related to avoiding `self-serving bias' in behavioral economics \cite{konow2003fairest}. Here, the goal is to understand the extent of which having `skin in the game' can impact human perception, and whether fairness axioms should be sensitive to such biases.
\textit{Would an envy-free agent consider the outcome fair if they were to observe it as a bystander?}

One approach is to use the judgement of the other individuals as a proxy for objectivity by measuring fairness in `the eyes of the others' \cite{shams2022fair}; thus, envy exists if sufficiently many agents agree that it should be the case (if they were in his shoes).
It is  a generalization of the unanimous envy condition proposed by \citet{van1995real}. 
An open question is, thus, whether individuals tend to value the views of others (and perhaps those closer to them) when assessing the fairness of outcomes.

\subsubsection{Agency \& Deliberation.}



Given the key role of `agency' in human perception, an intriguing questions arises on whether (and to what extent) distributed fair algorithms are perceived to be fairer compared to centralized algorithms (see the following discussion on procedural vs. distributive fairness).
%
The impact of agency through outcome control, i.e. the ability to adjust the outcomes, has shown to improve the perceived fairness among participants in resource allocation studies \cite{Lee19:Justice}. More importantly, even when allocations satisfy the fairness notion of EF1, the sheer `sense of control and agency' resulted in higher satisfaction even when modifications were performed as a group.
In voting, \citet{grandi2022voting} has recently shown that group deliberation in a repeated setting improves the social outcome where participants demonstrated a more optimistic behavior when given agency to deliberate.
Similar experiments in fair resource allocation illustrate that agency and group deliberation improve participants perception of fairness in comparison with outcomes prescribed by algorithms on Spliddit \cite{Lee17:Mediation}.
In the context of matching, a recent experimental study, similarly, illustrates the role of autonomy and agency in perceived fairness of matching algorithms \cite{konig2019fair}.

Thus, the following questions arise: \textit{How (and to what extent) involving participants in the decision-making process improves their perception of fairness? And how the perceived fairness relates to human cognitive biases such as the \textit{Ikea Effect} \cite{norton2012ikea}---tendency to associate higher value to products/outcomes one contribute---and cognitive dissonance, as observed by \citet{konow2003fairest}?}



\subsubsection{Motivation, Context, and Framing.}

Individuals' motivation for a task as well as the context (e.g. culture) impacts their actions and perceptions. A recent study on resource allocation \cite{lee18:understanding} suggests that the perceived fairness of algorithms is \textit{dependent on task characteristics}; and participants consider algorithmic solutions to be equally fair in mechanical tasks (e.g. processing data) compared to solutions suggested by humans, while they perceive algorithm-induced outcomes to be less fair for tasks that require human skills and subjective judgement.

Moreover, the sensitivity of fairness axioms to the well-known \textit{framing effect} \cite{tversky1985framing}--whether in evaluating outcomes or during the elicitation of preferences--could impact the choice of fairness notions in collective decision-making.
Framing effect often causes \textit{preference reversal} which states that a change in reference point may lead to a
change in preference (aka preference reversal) 
even if the values associated with possible outcomes remain
unchanged. \citet{kahneman1982judgment} showed that human behavior rarely adheres to a closely-related axiom, often known as the independence axiom (aka Independence of Irrelevant Alternative).

Thus, an immediate direction is studying how context and framing could impact the perception of fairness. The recent focus in studying allocation algorithms pertaining to goods (positively valued items) versus chores (negatively valued items) and mixture thereof, albeit necessary, is primarily theoretical. For example, axioms such as EF1 and EFX can easily be interpreted when distributing tasks by considering the removal of item from one's own bundle; yet, the perceived fairness of these solutions may be impacted by human's behavioral change when facing gains vs. losses---a seminal concept in \textit{prospect theory} \cite{kahneman1979prospect}.


\subsubsection{Distributive vs. Procedural Justice.}\label{sec:procedural}

The vast majority of research in fair division can be situated within \textit{distributive justice}\footnote{Distributive justice has a long history in philosophy within Rawls' theory of justice in distribution of social goods. We refer the readers to the seminal works of \citet{adams1963towards} and \citet{rawls1971theory}.}, which is primarily concerned with socially fair `outcomes'. However, the human judgement of decisions is often impacted by perceived fairness of the procedures. 
In other words, the interaction between distributive fairness and \textit{procedural justice} \cite{tyler2002procedural} in allocation of resources---particularly when dealing with approximate fairness axioms---requires an in-depth investigation.

Let us revisit the instance given in Example \ref{ex:runningexample}: both circled and underlined allocations satisfy EF1; however, the circled allocation may be perceived to be fairer as it is the outcome of a fair procedure (in this case the Round-Robin algorithm):  agents pick their favorite items among the remaining items one by one according to the $1, 2, 3, 1$ ordering.
More importantly, note that the underlined allocation is theoretically more desirable as it results in higher social welfare (it is simultaneously EF1 and Pareto optimal).

In the context of allocating a divisible resource, procedures that are envy-free are perceived to be fairer by participants than those that are proportional (but necessarily not envy-free) \cite{kyropoulou22:fair}. These findings imply that participants--all things equal--can correctly identify (and prefer) procedures that provide stronger fairness guarantees. Yet, the comparison between fair procedures and fair outcomes remains an interesting open problem. \textit{Are envy-free outcomes through centralized algorithms perceived to be fairer compared to fair procedures that only guarantee weaker notions of fairness (e.g. proportionality)?}



\section{Uncertainty and Temporal Effects} \label{sec:uncertainty}
Thus far we mainly highlighted challenges in devising and evaluating axioms of fairness in isolated, static, settings.
Yet, human judgement is often impacted by uncertainty of the outcomes or procedures as well as dynamic or temporal nature of parameters or decisions.


\subsubsection{Fair Lotteries.} Randomization has emerged as an alternative approach to approximate fairness notions such as EF1. The goal is often to achieve fairness ex-ante by allowing agents to participate in fair lotteries.
While recent efforts investigated theoretical boundaries of achieving ex-ante envy-freeness and ex post EF1 solutions \cite{freeman2020best,aziz2020simultaneously}, the interaction between the two approach and how they impact the perception of fairness remain mainly open.

\subsubsection{Sequential Fairness.}
When items need to be allocated sequentially, for example in charity donations, human perception may be affected by its lookahead reasoning, different discounting factors for future allocations, as well as availability bias.
Similarly, in problems involving repeated allocation (or reallocation) of resources, the following questions may arise: Which axioms of fairness can adequately capture human judgements? And how should these solutions be implemented to improve perceived fairness?

\subsubsection{Dynamic Settings.}
Imagine a scenario where new individuals arrive after an allocation decision is made. What modifications or redistribution of resources are considered socially acceptable (and fair)?
Similarly, when an agent leaves its items behind, how shall we redistribute its share among the rest of agents so as to maintain fairness? What constitutes a fair redistribution? Would the initial allocation's (un)fairness impact how the resdistribution should happen?
For example, consider an envy-free allocation. An agent leaves and the resources it held now needs to be redistributed. 
Suppose redistribution $A$ preserves envy-freeness, but another redistribution, say $B$, does not. Which redistribution is perceived to be fairer? 
What if the initial allocation is not envy-free?

\section{Verification of Fairness}

\subsection{Proof of Fairness}

A key criterion in adopting algorithmic approaches for solving societal problems is human perception. 
The majority of algorithms and procedures were not traditionally designed with human perception in mind.
In particular, fairness as distributive justice (aka, outcome fairness) is challenging to verify outside of academic and AI expert circles; we cannot expect from people the level of algorithmic literacy that is necessary for understanding and verifying solutions. 
Even if we do, it is unreasonable to presume sufficient cognitive (or computational) dedication to verify the outcomes generated by algorithms.
Note that even when dealing with AI-powered agents working on behalf of firms, institutions, or individuals, solutions (and their certificates) are ultimately interpreted by human stakeholders.
Thus, a pressing research agenda is how to provide a \textit{proof of fairness} that takes human perception into account.

On the philosophical front, proof of fairness is closely related to \textit{interactional justice} \cite{bies1986interactional}, that focuses on explanations that are provided to individuals to justify mechanisms that are used to ensure procedural fairness or delineate the reasons behind distribution of resources in a certain fashion.\footnote{Interactional justice has two components: \textit{interpersonal justice}, which focuses on treating individuals with dignity and respect by social planners or authorities, and \textit{informational justice}, which is focused on the explanation of outcomes or procedures \cite{greenberg1990organizational,greenberg1993social}. Here, we are primarily referring to the latter.}
These issues, at heart, require a thorough understanding of human perception, human-AI interaction, and algorithmic thinking. 

\subsection{Privacy-Preserving Verification}

Providing a proof of fairness to participants may have severe privacy ramifications. For example, verifying envy-freeness requires sharing information about the bundles allocated to all other agents. Thus, when sensitive information needs to be shared to ensure fairness (e.g. the availability of an agent, location information in facility location problems, and evaluating colleagues or team members in an organization), there is a need for properly balancing privacy concerns with the transparency required to verify fairness of an outcome.
A recent work on differential privacy in fair allocation of indivisible goods \cite{manurangsi2022differentially} show mainly negative results in achieving approximate fairness notions of envy-freeness and proportionality. 
An overarching question is \textit{how should we devise verification mechanisms that allow for reasonable and practical proof of fairness while ensuring some level of privacy?}

A promising direction is potentially focusing on information withholding approaches (as we discussed in Section~\ref{sec:transparency}). However, these approaches may in turn have negative impact on perceived fairness. 
Another practical approach for privacy-preserving fairness verification is to exploit randomization techniques. For instance, \textit{should we enable the individuals to randomly select (a limited number of) bundles to verify fairness (e.g. envy-freeness) instead of sharing information about other agents? How would this approach in return impact individuals' perception of fairness or more importantly their trust in algorithmic decision-making?}

\subsection{Explainable Fairness}

Taking an AI standpoint, there has been much interest in explainable AI within the machine learning community which primarily focuses on transparency in the mechanics of the machine learning algorithms \cite{burkart2021survey,Chakraborti2022contrastive}. We advocate for a broader sense of explainablity, that is, solutions must be explainable not only to experts and system designers but they should also provide justifications to the participating parties.

Explainability in mechanism design and voting has attracted some attention in recent years primarily from a theoretical point of view. 
Explainable mechanism design often has to deal with unique challenges (in contrast to the general explainable AI framework) due to complexities such as multi-objective norms, different stakeholders, and the range of the required justifications \cite{suryanarayana2022explainability}.
In the context of voting, recent works focus on providing justifications for a given decision through a set of norms and axioms that are `collectively accepted' by the agents along with arguments that explain the solutions \cite{boixel2020automated,nardi2022graph}. 
While the size of explanations can be bounded in certain domains of voting \cite{peters2020explainable}, how individuals measure and evaluate these explanations remains mainly unstudied. 
In addition, axioms of fairness rely on a variety of social norms and cognitive biases that could mutually interact with the framing, size, or the ordering of the explanations.


\subsection{Human-AI Interactions and Visualization}

A primary challenge in designing fair solutions is understanding how people and AI systems interact with one another. These interactions require a reasonable (and sufficient) methods to elicit complex preferences as well as communicating the outcomes (or potential outcomes) to the individuals.

\subsubsection{Preference Elicitation.}
In fair allocation of indivisible items, we often consider restricted preferences (e.g. additive valuations) to allow for compact representation of preferences over bundles.
While these restrictions have origins in practical applications, it is unreasonable to expect that individuals can assign \textit{precise} cardinal valuations to a large number of items. 
On the other hand, ordinal preferences may not contain sufficient information required for total extensions over outcomes.
%

Preference elicitation in systems involving human participants requires going beyond theoretical models that aim at minimizing the number of queries. Rather, it requires proper mechanics about how to (and to what extent) interact to prevent cognitive overload and maintain a balance between optimal elicitation and achieving socially desirable decisions.
%

\subsubsection{Visualizations.}
Visualization of the fair algorithms such as Spliddit \cite{shah2017spliddit} and MatchU.ai \cite{ferris2020matchu} could improve the algorithmic literacy of individuals. Such efforts have been shown to improve algorithmic thinking and providing  (limited) ability for comprehending certain axioms \cite{bao2023mind}. Yet, these visualizations are insufficient in providing proof of fairness that can then be simply verified by individuals.
Thus, the verification of distributive justice (outcome fairness) not only requires fairness axioms to be coherent with human values, but also calls for equipping individuals with smooth visual interactions to check and ultimately accept the solutions.

\section{Conclusion}

Fairness issues in AI systems arise from a variety of axiomatic, algorithmic, and epistemic assumptions. These challenges require a precise and in-depth interdisciplinary discussions to align fairness axioms with human value judgements, evaluate the perceived fairness of common fairness notions, and develop new axioms of fairness. Novel algorithmic approaches should enable individuals to verify the fairness of the solutions by providing explanations for the procedures, enabling interactions with the algorithms, and providing a proof of fairness while preserving privacy of the individuals.
From the technical  perspective, new algorithms should be designed to adhere to the new fairness axioms that are aligned with human values. Computational hardness may not necessarily be an obstacle if the new axioms are shown to not rely on full information. In this vein, theoretical research may shift focus from approximate methods to context-aware solutions.

Incorporating human perception in collective decision-making is an extremely challenging task. A long-term successful plan must draw from a broad array of disciplines including AI and computing research to experimental economics and behavioral psychology.
Our aim in this paper was to paint a broad agenda for future research in collective decision-making. The hope is to enrich our understanding of human judgement and develop algorithmic solutions that are responsive (or at least aware) of human perception and ultimately establish new axioms of fairness for future research.

\section*{Acknowledgments}
We are grateful to the anonymous reviewers for their insightful comments. We would like to thank Tomasz W{\k{a}}s for helpful feedback on the paper. 
This work is supported by NSF IIS grants \#2144413 (CAREER) and \#2107173.

\bibliography{fairref,pubs,mixedref,ref_MMS}


\end{document}